\documentclass{article}

\usepackage{arxiv}

\usepackage{wrapfig}
\usepackage[utf8]{inputenc}
\usepackage[T1]{fontenc}
\usepackage{amsmath}
\usepackage{amsfonts}
\usepackage{booktabs}
\usepackage{graphicx}
\usepackage{microtype}
\usepackage[authoryear,round]{natbib}
\usepackage{xcolor}
\definecolor{HLCiteBlue}{HTML}{0069AA}
\definecolor{HLRefBrown}{HTML}{8C4628}
\usepackage{url}
\usepackage{algorithm}
\usepackage{algpseudocode}

\usepackage{array}     
\usepackage{tabularx}  
\usepackage{tikz}
\usetikzlibrary{arrows,arrows.meta,calc,positioning}
\usepackage{hyperref}
\hypersetup{
  colorlinks=true,
  citecolor=HLCiteBlue,
  linkcolor=HLRefBrown,
  urlcolor=HLCiteBlue
}

\newcommand{\clip}{\operatorname{clip}}
\newcommand{\mean}{\operatorname{mean}}

\newcommand{\interp}{\operatorname{interp}}
\newcommand{\sgn}{\operatorname{sgn}}

\graphicspath{{plots/}{plots/per_env/}{plots/attached_clean/}}

\newcommand{\env}[1]{\texttt{#1}}

\newcommand{\score}[1]{\ensuremath{#1}}

\title{Heuristic Learning for Active Flow Control Using Coding Agents}

\author{
  Paul Garnier \\
  Mines Paris - PSL University \\
  Centre for Material Forming (CEMEF) \\
  CNRS \\
  \texttt{paul.garnier@minesparis.psl.eu} \\
  \And
  Jonathan Viquerat \\
  Independent Researcher \\
  \texttt{jonathanviquerat@gmail.com} \\
  \And
  Elie Hachem \\
  Mines Paris - PSL University \\
  Centre for Material Forming (CEMEF) \\
  CNRS \\
  \texttt{elie.hachem@minesparis.psl.eu} \\
}

\begin{document}
\maketitle

\begin{abstract}

Active flow control involves nonlinear dynamics, partial observations, and computationally expensive simulations, making controller design particularly challenging. Deep reinforcement learning (DRL) has emerged as a powerful framework for such problems, but its success typically relies on large numbers of simulator interactions and produces neural-network policies whose decision process often remains difficult to interpret. In this work, we investigate a different paradigm: instead of optimizing neural-network parameters, we use modern coding agents to search directly for explicit executable feedback laws. We introduce a constrained heuristic-learning protocol in which an agent iteratively proposes, evaluates, and revises controller implementations while interacting exclusively through the public benchmark interface. The proposed framework is evaluated on 13 active flow-control benchmarks spanning one, two, and three-dimensional problems and compared against the strongest available DRL baselines under identical simulation budgets. The discovered heuristic controllers match or outperform the best DRL policy in 10 of the 13 environments while remaining compact, interpretable, and directly inspectable. Beyond aggregate performance, the resulting controllers reveal physically meaningful feedback mechanisms, transfer successfully across more challenging configurations, and remain competitive under varying Reynolds and Rayleigh numbers, actuator counts, and observation sparsity. These results suggest that heuristic learning through coding agents constitutes a credible and complementary alternative to conventional reinforcement learning, combining competitive performance with physically interpretable controller representations. Prompts and source code are available at \url{https://github.com/DonsetPG/fluid-heuristic-learning}.

\end{abstract}


\section{Introduction}

Active flow control remains one of the central challenges in computational fluid dynamics. The objective is to determine a feedback strategy capable of modifying the evolution of a nonlinear flow using only partial observations and a limited set of admissible actuators. Such control problems are characterized by high-dimensional state spaces, nonlinear dynamics, delayed responses, and expensive numerical simulations, making them particularly difficult to solve with classical optimization techniques. Over the last decade, deep reinforcement learning (DRL) has therefore emerged as a natural framework for flow control, as it does not require an adjoint solver, the differentiability of the complete simulation chain, or a prescribed functional form for the controller. Instead, a neural policy interacts with the environment by (i) receiving a reduced set of observations from the full flow state, (ii) processing them into one or several actions, (iii) applying these actions through the admissible control channels, and (iv) receiving a scalar reward that measures the quality of its behavior while optimizing a long-term estimate of this reward.

This simple interaction loop has led to a rapid sequence of successes across a broad range of fluid mechanics applications, including stabilization and drag reduction in the cylinder wake using synthetic jets \citep{rabault2019active}, control of falling-film instabilities \citep{belus2019exploiting}, thermal control in Rayleigh--B\'enard convection \citep{beintema2020controlling, hachem2021conjugate}, collective swimming strategies \citep{novati2017synchronisation,verma2018efficient}, direct aerodynamic shape optimization \citep{viquerat2021shape}, and experimental bluff-body flow control \citep{fan2020reinforcement}. These studies demonstrated that DRL can discover highly non-trivial feedback strategies directly from interaction with the environment, and in several cases recover controllers whose underlying physical mechanisms can be interpreted a posteriori.

Despite these successes, the flexibility of DRL comes at a cost. Training a controller typically requires a very large number of rollouts, each involving the solution of a CFD problem over hundreds or thousands of time steps. Although this computational burden can be partially alleviated through parallel environment sampling \citep{rabault2019accelerating,viquerat2023parallel}, it remains one of the main bottlenecks of the approach. Furthermore, the resulting controller is generally represented by a neural network. While such policies may achieve excellent performance, their internal decision process often remains difficult to interpret, analyze, modify, or transfer to related configurations. Reproducibility also remains an important concern: relatively small variations in reward formulation, observation selection, action smoothing, random initialization, or implementation details can affect the final learned policy \citep{andrychowicz2020what,viquerat2022review}. These limitations do not question the relevance of DRL for flow control; rather, they motivate a complementary scientific question: can explicit, human-readable control laws be discovered automatically while remaining competitive with state-of-the-art reinforcement learning?

Recent advances in large language models (LLMs) suggest that such a possibility may now be within reach. Since the introduction of neural attention mechanisms \citep{bahdanau2015neural} and the Transformer architecture \citep{vaswani2017attention}, LLMs have rapidly evolved from language-processing systems into increasingly capable reasoning and programming assistants \citep{chen2021evaluating,jimenez2024swebench}. More recently, coding agents such as Claude Code \citep{anthropic2026claudecode} and Codex \citep{openai2025codex} have demonstrated the ability to execute long-horizon workflows that may span hours or even days \citep{openai2026codexapp,anthropic2026longrunningclaude}. Rather than optimizing millions of neural-network parameters through gradient descent, these systems make it possible to search directly in the space of executable programs.

This shift opens a fundamentally different perspective for flow control. Instead of learning an implicit policy encoded in neural-network weights, a coding agent can iteratively propose explicit feedback laws, evaluate them through numerical simulations, analyze their performance, and refine the corresponding controller based on observed successes and failures. The resulting search process no longer explores a parameter space, but a space of interpretable algorithms composed of filters, delays, feedback gains, state machines, or other physically meaningful operations. Such explicit controllers can be inspected, modified, transferred, and analyzed by fluid dynamicists, potentially restoring an important level of interpretability that is often sacrificed in purely neural approaches.

The present work investigates whether this new paradigm can serve as a practical alternative for active flow control. More specifically, we study whether modern coding agents can autonomously discover explicit feedback controllers that remain competitive with state-of-the-art DRL while respecting exactly the same observations, actions, rewards, and interaction budget. To ensure a fair comparison, the agents are constrained to interact exclusively through the public benchmark interface and are prohibited from accessing hidden simulator information or modifying the environment itself.

This paper makes four main contributions:

\begin{enumerate}
\item We introduce a reproducible heuristic-learning protocol in which coding agents autonomously search for explicit feedback controllers under strict public-interface constraints and invalid-policy filtering.
\item We evaluate this protocol on thirteen fluid-dynamics control benchmarks using multiple coding agents, and compare the resulting controllers against the strongest available DRL baselines.
\item We show that automatically discovered heuristic controllers frequently match or outperform DRL while remaining compact, explicit, and physically interpretable.
\item We investigate the robustness and generalization of the proposed approach through transfer learning, variable-dimensional action spaces, and sparse observation settings.
\end{enumerate}

The remainder of the paper is organized as follows. Section~\ref{sec:related-work} reviews previous work on DRL for flow control, benchmark suites, and program synthesis approaches. Section~\ref{sec:problem} introduces the control formulation, while Section~\ref{sec:hl} presents the proposed heuristic-learning protocol. Section~\ref{sec:experiments} describes the benchmark suite and the experimental setup. Section~\ref{sec:results} reports the main results together with additional analyses, and Section~\ref{sec:policy-anatomy} examines representative controllers to better understand the physical strategies discovered by the coding agents.

\section{Related Work}
\label{sec:related-work}

\subsection{Deep reinforcement learning for flow control}

Deep reinforcement learning has progressively established itself as one of the principal paradigms for active flow control by reformulating the control problem as a sequential interaction between an agent and its environment. In this framework, the environment may consist of a numerical solver, a reduced-order model, or a physical experiment. The agent receives partial observations of the flow, selects control actions, modifies boundary conditions, source terms, body motion, geometry, or material parameters, and finally receives a reward that quantifies the quality of the chosen strategy.

One of the earliest and most influential demonstrations was the stabilization of the two-dimensional cylinder wake using synthetic jets, where a PPO policy successfully reduced drag while suppressing the von Kármán vortex street \citep{rabault2019active}. Beyond its physical results, this work introduced a reusable CFD--DRL coupling that later benefited from efficient parallel environment sampling \citep{rabault2019accelerating}. Since then, similar formulations have been extended to different Reynolds numbers, alternative actuation mechanisms, weakly turbulent regimes, sparse and noisy observations, and even experimental setups \citep{tang2020robust,ren2021applying,fan2020reinforcement}. More broadly, DRL has been successfully applied to falling-film stabilization, conjugate heat transfer, swimming strategies, microfluidics, turbulence-related problems, aerodynamic shape optimization, and many other fluid-mechanics applications \citep{belus2019exploiting,beintema2020controlling,garnier2021review,hachem2024quenching}.

Although the targeted physical systems differ considerably, several methodological choices have become common throughout the literature. Most studies rely on model-free actor--critic algorithms, with PPO being the dominant on-policy baseline, while TD3 and related off-policy methods are occasionally considered for comparison \citep{schulman2017ppo,fujimoto2018addressing,viquerat2022review}. Observations are typically restricted to a limited number of probes or physically meaningful quantities extracted from the full flow field, whereas rewards are designed as low-dimensional proxies of engineering objectives such as drag, lift, heat transfer, concentration variance, or interface deformation. These design choices strongly influence the resulting controller. Sensor placement, reward shaping, control frequency, action smoothing, random initialization, and implementation details all affect learning dynamics and final performance. It is now well recognized that two implementations of the same DRL algorithm may produce significantly different results on an identical benchmark, motivating recent efforts toward standardized environments, open-source implementations, and reproducible evaluation protocols \citep{andrychowicz2020what}.

In the present work, DRL serves exclusively as a reference methodology. We deliberately preserve the same environments, observations, actions, rewards, and evaluation protocols, but replace neural-network optimization by the search for explicit executable controllers. This allows the comparison to focus solely on the representation and discovery of the control law rather than on differences in the underlying control problem.

\subsection{Benchmark suites}

Benchmarking plays a particularly important role in reinforcement learning, where algorithmic performance is often highly dependent on the considered environments. Classical benchmarks such as Atari, MuJoCo, and OpenAI Gym have been instrumental in the development of modern RL algorithms, yet they do not reproduce many of the constraints encountered in computational fluid dynamics, including expensive numerical rollouts or partially observed spatio-temporal fields\citep{bellemare2013arcade,todorov2012mujoco,brockman2016openai}. Consequently, improvements demonstrated on generic RL benchmarks do not necessarily translate into advances for flow-control applications.

BEACON was introduced to provide a lightweight common ground for this purpose
\citep{viquerat2024beacon}. It contains seven self-contained one- and
two-dimensional environments with different physics, observation spaces, action
spaces, control types, and CPU requirements. The design constraints are
practical: Python implementation, Gym-like interface, low enough cost to allow
training on a workstation, and reference curves for standard DRL algorithms.
The suite therefore plays the role of a first filter. It is broad enough to
include problems such as falling films, thermal convection, mixing or sloshing, yet computationally cheap enough to quickly screen algorithmic issues and inefficiencies before moving to more CPU-expensive environments.

FluidGym \citep{becktepe2026fluidgym} targets larger active-flow-control problems in a single PyTorch stack, with
differentiable environments, GPU acceleration, standardized evaluation
protocols, and baselines for modern RL algorithms. This makes it closer to the
large-scale setting in which poor sample efficiency and brittle controller
design become limiting factors. In the following experiments, we use BEACON for
fast and diverse iteration and FluidGym to test whether the same
heuristic-learning protocol still behaves sensibly when each evaluation is
substantially more costly.

\subsection{Programmatic policies and coding-agent search}

The objective of the present work is fundamentally different from conventional reinforcement learning. Rather than optimizing the parameters of a neural policy, we seek to automatically construct explicit executable controllers through an iterative program-synthesis process. Our approach is therefore more closely related to recent work on language-model optimization and autonomous code generation than to classical policy-gradient methods.

A first line of research treats natural language as the optimization medium itself. Methods such as Feedback Descent, GEPA, ProTeGi, Self-Refine, TextGrad, and OPRO iteratively improve candidate solutions by storing critiques, textual gradients, pairwise preferences, rollout summaries, or optimization histories directly in textual form \citep{lee2025feedback,agrawal2025gepa,pryzant2023protegi,madaan2023self,yuksekgonul2024textgradautomaticdifferentiationtext,yang2023large}. Rather than updating numerical parameters through gradient descent, these approaches progressively refine future proposals by accumulating structured textual experience. Our setting introduces an additional requirement: every generated proposal must immediately become a valid flow controller, interact exclusively through the official environment interface, satisfy all actuator constraints, survive deterministic rollout evaluation, and remain sufficiently transparent to be inspected by a fluid dynamicist.

A second family of methods considers executable programs as the optimization object. Systems such as OpenEvolve, AlphaEvolve, and FunSearch repeatedly modify source code, execute the resulting programs, evaluate their performance, and preserve only successful variants \citep{openevolve,novikov2025alphaevolve,lehman2022evolutionlargemodels,romera2024mathematical}. More recently, ADAS and AFlow extended this philosophy beyond individual programs by allowing entire agent architectures or workflow graphs to evolve during optimization \citep{hu2025automated,zhang2025aflowautomatingagenticworkflow}. These approaches are structurally the closest to the framework proposed here. In all cases, language models propose modifications, execution provides objective feedback, and subsequent revisions are guided by observed successes and failures. Our contribution specializes this general paradigm to feedback control in computational fluid dynamics, where generated programs must additionally satisfy strict physical and interface constraints.

Finally, our work is also related to broader research on memory, meta-learning, and adaptive optimization. MemEvolve and meta-learned agentic memory investigate how an autonomous agent should retain, retrieve, and exploit previous experiences throughout long optimization processes \citep{zhang2025memevolve,Xiong2026LearningTC}. Earlier work on self-referential learning systems, learning-to-learn, learned optimizers, MAML, prototypical networks, and in-context learning similarly addresses the broader question of how previous experience can accelerate future adaptation \citep{schmidhuber1993self,thrun1998learning,andrychowicz2016learning,finn2017maml,snell2017prototypical,akyürek2023learningalgorithmincontextlearning}. Unlike these approaches, however, the present work does not adapt model parameters. Instead, it searches directly in the space of explicit control programs, allowing every candidate policy to remain executable, inspectable, and physically interpretable throughout the optimization process.

\section{Problem Formulation}
\label{sec:problem}

\begin{figure}[t]
\centering
\resizebox{0.98\linewidth}{!}{\definecolor{hlink}{RGB}{28,34,42}
\definecolor{hlmuted}{RGB}{95,105,117}
\definecolor{hlgrid}{RGB}{223,229,236}
\definecolor{hlblue}{RGB}{31,119,180}
\definecolor{hlteal}{RGB}{0,140,135}
\definecolor{hlamber}{RGB}{198,131,28}
\definecolor{hlbluefill}{RGB}{235,244,251}
\definecolor{hltealfill}{RGB}{231,246,244}
\definecolor{hlamberfill}{RGB}{252,244,231}
\definecolor{hlgrayfill}{RGB}{247,249,251}

\newcommand{\HLLoopTitle}[1]{\textcolor{hlmuted}{\scriptsize #1}}

\newcommand{\HLRLFigure}{%
\begin{tikzpicture}[
  >=Latex,
  font=\scriptsize,
  node distance=1.15cm and 1.35cm,
  box/.style={
    rectangle,
    rounded corners=2pt,
    draw=hlmuted,
    line width=0.55pt,
    minimum width=2.25cm,
    minimum height=0.82cm,
    align=center,
    inner sep=3pt,
    fill=hlgrayfill
  },
  data/.style={font=\scriptsize, text=hlmuted, inner sep=1pt, fill=white},
  arrow/.style={->, line width=0.65pt, draw=hlmuted},
  update/.style={->, line width=0.75pt, draw=hlamber}
]
  \node[box, fill=hlbluefill] (policy) {Policy network\\$\pi_\phi(a\mid s)$};
  \node[box, right=2.65cm of policy, fill=hltealfill] (env) {Environment\\$s_t \rightarrow s_{t+1}$};
  \node[box, below=1.28cm of $(policy)!0.5!(env)$, fill=hlamberfill] (learn) {Optimizer\\updates weights $\phi$};

  \draw[arrow] (policy) -- node[data, above] {$a_t$} (env);
  \draw[arrow] (env.south west) to[out=-145,in=20] node[data, right] {$s_{t+1}, r_t$} (learn.east);
  \draw[update] (learn.west) to[out=160,in=-95] node[data, left] {gradient} (policy.south);
\end{tikzpicture}%
}

\newcommand{\HLHeuristicFigure}{%
\begin{tikzpicture}[
  >=Latex,
  font=\scriptsize,
  node distance=1.0cm and 1.18cm,
  box/.style={
    rectangle,
    rounded corners=2pt,
    draw=hlmuted,
    line width=0.55pt,
    minimum width=2.18cm,
    minimum height=0.78cm,
    align=center,
    inner sep=3pt,
    fill=hlgrayfill
  },
  data/.style={font=\scriptsize, text=hlmuted, inner sep=1pt, fill=white},
  arrow/.style={->, line width=0.65pt, draw=hlmuted},
  codearrow/.style={->, line width=0.78pt, draw=hlblue},
  searcharrow/.style={->, line width=0.72pt, draw=hlamber}
]
  \node[box, fill=hlbluefill] (code) {Readable controller\\$a_t=h_\theta(w_t)$};
  \node[box, right=2.55cm of code, fill=hltealfill] (env) {Simulator\\flow rollout};
  \node[box, below=1.18cm of env, fill=hlgrayfill] (score) {Traces and score\\$r_t, C_D, E, ...$};
  \node[box, below=1.18cm of code, fill=hlamberfill] (edit) {Code edit\\rules, gains, features};

  \draw[codearrow] (code) -- node[data, above] {actions} (env);
  \draw[arrow] (env) -- node[data, right] {evidence} (score);
  \draw[searcharrow] (score) -- node[data, below] {diagnose} (edit);
  \draw[codearrow] (edit) -- node[data, left] {new $\theta$} (code);
\end{tikzpicture}%
}

\begin{tabular}{cc}
\HLLoopTitle{(a) Reinforcement learning} & \HLLoopTitle{(b) Heuristic learning}\\[3pt]
\HLRLFigure & \HLHeuristicFigure
\end{tabular}}
\caption{Comparison between the standard reinforcement-learning interaction loop and heuristic learning, where the agent revises an explicit rule-based policy from rollout outcomes.}
\label{fig:rl}
\end{figure}
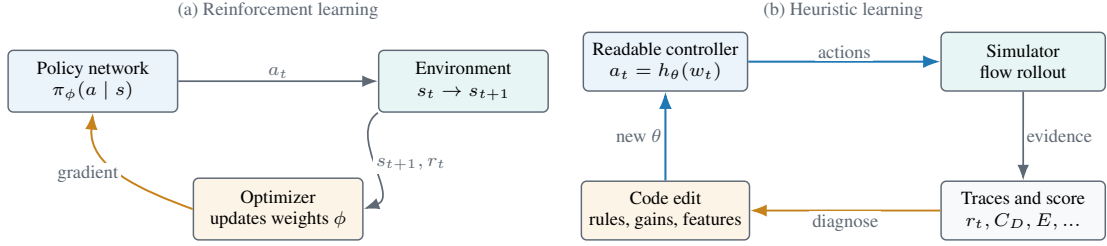

The control problems considered throughout this work follow the standard reinforcement-learning formulation. An agent repeatedly interacts with a controlled dynamical system, observes part of its state, applies an admissible control action, and receives a scalar reward measuring the quality of the resulting behavior. In the present context, the environment corresponds to a numerical fluid-dynamics simulator, while the agent represents the controller to be optimized. The interaction loop is illustrated in \autoref{fig:rl}.

Formally, the problem is modeled as a Markov decision process. At each iteration, the following sequence of operations is performed:

\begin{enumerate}
    \item At iteration $t$, the environment is in state $s_t \in \mathcal{S}$, where $\mathcal{S}$ denotes the state space.
    \item The controller receives an observation $w_t$, which may represent only a partial observation of $s_t$, and computes an action $a_t \in \mathcal{A}$, where $\mathcal{A}$ is the admissible action space.
    \item The environment evolves according to the controlled dynamics, producing the next state $s_{t+1}$.
    \item The environment returns both the next observation $w_{t+1}$ and a scalar reward $r_t \in \mathcal{R}$ quantifying the quality of the selected action.
\end{enumerate}

Repeating this interaction generates a trajectory

\[
\tau = \left(s_0,a_0,s_1,a_1,\ldots\right),
\]

which terminates once the episode reaches its prescribed horizon. The objective of reinforcement learning is to determine a policy $\pi$ maximizing the expected discounted cumulative reward,

\begin{equation}
\label{eq:discounted_cumulative_reward}
J(\pi)
=
\mathbb{E}_{\tau\sim\pi}
\left[
\sum_{t=0}^{T}
\gamma^t r_t
\right],
\end{equation}

where $T$ denotes the episode length and $\gamma\in[0,1]$ is the discount factor controlling the relative importance of immediate and future rewards.

Modern DRL algorithms differ primarily in the manner by which this optimization problem is solved. On-policy approaches update the controller exclusively from trajectories generated by the current policy, whereas off-policy methods additionally reuse previously collected experience. Throughout this work, the proposed methodology is compared against two widely adopted representatives of these families: Proximal Policy Optimization (PPO), a robust on-policy algorithm based on constrained policy updates, and Soft Actor-Critic (SAC), an off-policy algorithm combining actor--critic learning with entropy regularization to improve exploration and sample efficiency.

Importantly, the optimization objective itself remains unchanged in the present work. The observations, actions, rewards, simulators, and evaluation protocols are exactly the same as those used by conventional DRL. The only difference lies in the representation of the controller. Instead of optimizing the parameters of a neural-network policy, we search directly for an explicit executable control law that maps the publicly available observations to admissible actions. Consequently, the comparison isolates the effect of the controller representation itself while preserving an identical control problem and evaluation criterion. The following section introduces the proposed heuristic-learning framework used to perform this search.

\section{Heuristic-Learning Protocol}
\label{sec:hl}

The objective of the proposed heuristic-learning framework is to replace the optimization of neural-network parameters by the optimization of explicit executable control laws. Rather than updating a policy through gradient descent, a coding agent iteratively designs, evaluates, and revises controller implementations based solely on their observed performance. Throughout this process, the simulator, reward function, environment dynamics, and control interface remain fixed. The agent is only allowed to modify the controller source code. Consequently, the optimization problem is identical to that encountered in DRL; only the representation of the controller and the search procedure differ. The complete protocol is summarized in Algorithm~\ref{alg:heuristic-learning}.

The restriction to explicit heuristics is a deliberate experimental choice rather than a limitation of the underlying coding agents. Modern coding agents are capable of generating substantially more complex software, including neural-network implementations. Such controllers are intentionally excluded from the present study in order to preserve interpretability, reproducibility, and direct inspection of the resulting control strategies.

Each candidate controller is represented by an executable program $c$ together with an explicit internal memory $m_t$. At every control step, the controller receives only the information publicly exposed by the benchmark environment and computes an admissible action according to

\[
a_t=\pi_c(o_{\le t},r_{<t},m_t),
\qquad
m_{t+1}=U_c(m_t,o_t,r_t,a_t),
\]

where $o_t$ denotes the public observation, $r_t$ the reward returned by the environment, $a_t$ the action applied through the official control interface, $\pi_c$ the controller implemented by program $c$, and $U_c$ its explicit memory-update rule. Histories such as $o_{\le t}$ and $r_{<t}$ contain only information available before the current action is selected.

\begin{wrapfigure}{r}{0.62\textwidth}
    \centering
    \vspace{-1.5em}
    \includegraphics[width=0.60\textwidth]{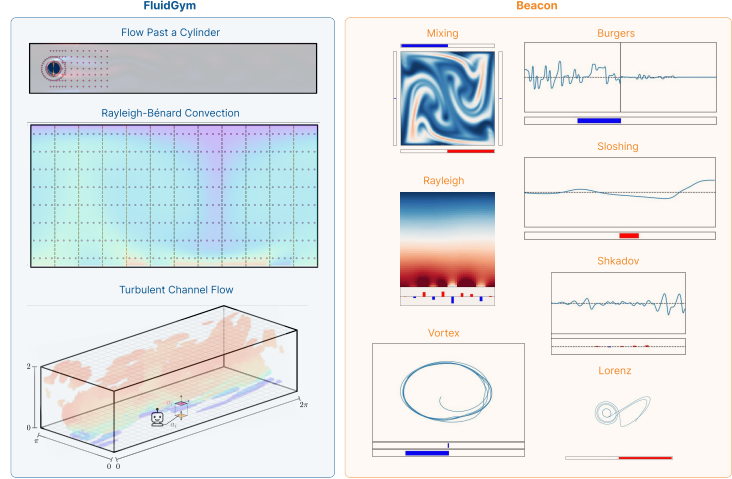}
    \caption{Presentation of the different environments. Abstract schematic on the left reproduced from \cite{becktepe2026fluidgym}.}
    \label{fig:experimental-setup}
    \vspace{-1.0em}
\end{wrapfigure}

The controller memory is itself subject to the same information constraints. It may store quantities derived exclusively from publicly available observations, including moving averages, delayed buffers, filtered signals, or previously observed states, but never hidden simulator variables or future information. For an environment family $\mathcal{E}$, a seed set $\mathcal{S}$, and a benchmark score $J$, the objective of the search is therefore

\[
c^\star
\in
\arg\max_{c\in\mathcal{C}_{\mathrm{valid}}}
\mathbb{E}_{e\sim\mathcal{E},\,s\sim\mathcal{S}}
\left[
J(c;e,s)
\right].
\]

The coding agent $A$ performs this optimization by repeatedly modifying controller source code according to

\[
c_{k+1}
=
A(c_k,\tau_k,J(c_k),\mathcal{L}_k),
\]

where $\tau_k$ denotes the rollout trace associated with controller $c_k$, and $\mathcal{L}_k$ is a public search ledger recording all information accumulated during previous evaluations. The ledger contains controller scores, action trajectories, rollout diagnostics, plots, physical metrics, ablation studies, failed hypotheses, transferred controllers, and previously identified invalid solutions. Unlike gradient-based optimization, the search therefore progresses through explicit reasoning over accumulated experimental evidence.

Each optimization cycle follows five successive stages. First, the agent inspects the public benchmark interface, including the observation structure, admissible action range, reward definition, reset procedure, diagnostics, and episode horizon. Second, it proposes a compact executable controller, for example a feedback law, delayed filter, finite-state machine, spatial projection, phase scheduler, or hybrid combination of these components. Although larger implementations are technically possible, only explicit train-free controllers are retained in the present study. Third, the controller is evaluated through deterministic rollouts, during which scores, diagnostics, actions, failures, and consumed simulation steps are recorded. Fourth, these observations are used to revise controller parameters or structure, including gains, clipping, delays, memory variables, spatial aggregation, or logical conditions. Finally, the surviving controllers are simplified through successive ablations before being validated on replay trajectories, unseen random seeds, longer horizons, transferred environments, or dedicated stress tests. An overview of the complete optimization cycle is provided in Table~\ref{tab:protocol}.

\begin{algorithm}[t]
\caption{Heuristic-learning loop over executable controller code}
\label{alg:heuristic-learning}
\begin{algorithmic}[1]
\State \textbf{Input:} environments $\mathcal{E}$, public interfaces $\mathcal{I}$, scoring function $J$, coding agent $A$, simulator budget $B$
\State \textbf{Initialize:} controller set $\mathcal{C}$ with no-control, simple feedback, or transferred seed policies
\State \textbf{Initialize:} public ledger $\mathcal{L} \leftarrow \emptyset$

\For{$c \in \mathcal{C}$}
    \If{$c$ passes static validity checks}
        \State $(J(c), \tau_c) \leftarrow \textsc{Evaluate}(c,\mathcal{E},\mathcal{I})$
        \State $\mathcal{L} \leftarrow \mathcal{L} \cup \{c,J(c),\tau_c,\textsc{Validity}(c)\}$
    \Else
        \State $\mathcal{L} \leftarrow \mathcal{L} \cup \{c,\textsc{Invalid}\}$
    \EndIf
\EndFor

\While{simulator budget $B$ remains}
    \State $A$ reads the public ledger $\mathcal{L}$
    \State $A$ proposes candidate controllers $\{c_1,\ldots,c_k\}$

    \For{$c_i$ in $\{c_1,\ldots,c_k\}$}
        \If{$c_i$ passes public-interface and source-code checks}
            \State $(J(c_i), \tau_i) \leftarrow \textsc{Evaluate}(c_i,\mathcal{E},\mathcal{I})$
            \State $\mathcal{L} \leftarrow \mathcal{L} \cup \{c_i,J(c_i),\tau_i,\textsc{Validity}(c_i)\}$
        \Else
            \State $\mathcal{L} \leftarrow \mathcal{L} \cup \{c_i,\textsc{Invalid}\}$
        \EndIf
    \EndFor
\EndWhile
\end{algorithmic}
\end{algorithm}

\begin{table}[t]
\centering
\small
\caption{Heuristic-learning loop used in the experiments. Prompts and run commands are provided in the supplementary material.}
\label{tab:protocol}
\begin{tabular}{p{0.18\linewidth}p{0.72\linewidth}}
\toprule
Stage & Role \\
\midrule
Interface probe & Inspect observation shape, action bounds, reward scale, reset behavior, public diagnostics, and episode horizon. \\
Policy proposal & Write a compact executable controller such as a feedback law, delayed filter, finite-state machine, spatial projection, schedule, or hybrid strategy. \\
Evaluation & Execute fixed-seed rollouts and record scores, physical diagnostics, action traces, and failure cases. \\
Early rejection & Interrupt weak candidate evaluations once public diagnostics indicate poor performance while accounting for all consumed simulation steps. \\
Revision & Update controller parameters or structure using the accumulated rollout evidence. \\
Simplification & Remove unnecessary branches through ablation while preserving interpretability. \\
Validation & Evaluate the surviving controllers on held-out seeds, longer horizons, transferred environments, or replay diagnostics. \\
\bottomrule
\end{tabular}
\end{table}

\subsection{Explicit constraints}

To ensure a fair comparison with DRL, the heuristic-learning framework is subject to three explicit protocol constraints.

\paragraph{Search budget accounting.}

Each heuristic search receives exactly the same total simulation budget as the corresponding DRL experiment. Whenever the coding agent terminates an evaluation early because a controller is clearly ineffective, the consumed simulation steps are nevertheless fully charged against the available budget. Consequently, heuristic learning may evaluate more candidate controllers than DRL, but never benefits from additional interactions with the environment.

\paragraph{Prompt-induced protocol violations.}

The coding agent is treated as an experimental operator whose actions must remain strictly confined to the public benchmark interface. Any attempt to modify the simulator, clone the environment for look-ahead evaluation, alter the reward definition, access hidden simulator variables, or bypass the official actuator interface is considered invalid and discarded before evaluation.

\paragraph{Validity constraints.}

A controller is considered valid only if it satisfies all of the following conditions:

\begin{itemize}
\item it relies exclusively on publicly available observations, rewards, and admissible actions;
\item it interacts with the environment solely through the official control interface;
\item it neither accesses nor modifies hidden simulator state;
\item it neither trains nor loads opaque learned policies, including pretrained neural-network checkpoints. Explicit numerical algorithms, filters, or analytical controllers are permitted provided that they operate exclusively on public information and remain fully inspectable;
\item it leaves the benchmark unchanged by preserving the original reward function, episode horizon, and environment dynamics throughout evaluation.
\end{itemize}

\section{Experimental Setup}
\label{sec:experiments}

The proposed heuristic-learning framework is evaluated on the benchmark suites provided by FluidGym \citep{becktepe2026fluidgym} and BEACON \citep{viquerat2024beacon}. These environments provide standardized flow-control problems together with publicly available DRL baselines, allowing a direct comparison between conventional reinforcement learning and the proposed program-synthesis approach. All DRL reference results were reproduced from the official open-source implementations%
\footnote{\url{https://github.com/safe-autonomous-systems/fluidgym} and
\url{https://github.com/jviquerat/beacon}} to ensure consistency of the reported comparisons.

The heuristic-learning experiments were conducted using several state-of-the-art coding agents in order to evaluate the robustness of the proposed protocol across different underlying language models. Unless stated otherwise, GPT-5.5 operating through Codex with \texttt{xhigh} reasoning effort \citep{openai2026gpt55,openai2026gpt55docs,openai2026codexmodels} is used as the primary agent and is referred to as \textit{Codex}. Additional experiments were performed using Claude Opus 4.7 with \texttt{xhigh} reasoning effort \citep{anthropic2026opus47,anthropic2026claudecodeModelConfig} (denoted \textit{Claude}), Fable 5 with \texttt{xhigh} reasoning effort \citep{anthropic2026fable} (denoted \textit{Fable}), and Gemini 3 Pro \citep{google2025gemini3,google2025gemini3cli} (denoted \textit{Gemini}). The corresponding command-line interfaces were Codex CLI \citep{openai2026codexcli}, Claude Code \citep{anthropic2026claudecode}, and Gemini CLI \citep{google2026geminicli}.

To ensure a fair comparison with DRL, every coding agent was allocated exactly the same simulation budget as the corresponding reinforcement-learning baseline, measured in total environment interactions. Consequently, all reported differences arise from the controller representation and search strategy rather than from additional access to the simulator.

The complete catalogue of benchmark environments is presented in Appendix~\ref{app:environments} and summarized in \autoref{tab:env-suite-details}.

\section{Main Results}
\label{sec:results}

We first evaluate the proposed heuristic-learning framework on the complete benchmark suite comprising 13 flow-control environments. Across all cases, the search successfully discovers a non-trivial explicit controller. We begin by comparing the best heuristic obtained on each benchmark with the strongest available DRL baseline, before analyzing the optimization trajectories and convergence behavior.

\begin{figure}[t]
    \centering
    \vspace{-1.5em}
    \includegraphics[width=0.60\textwidth]{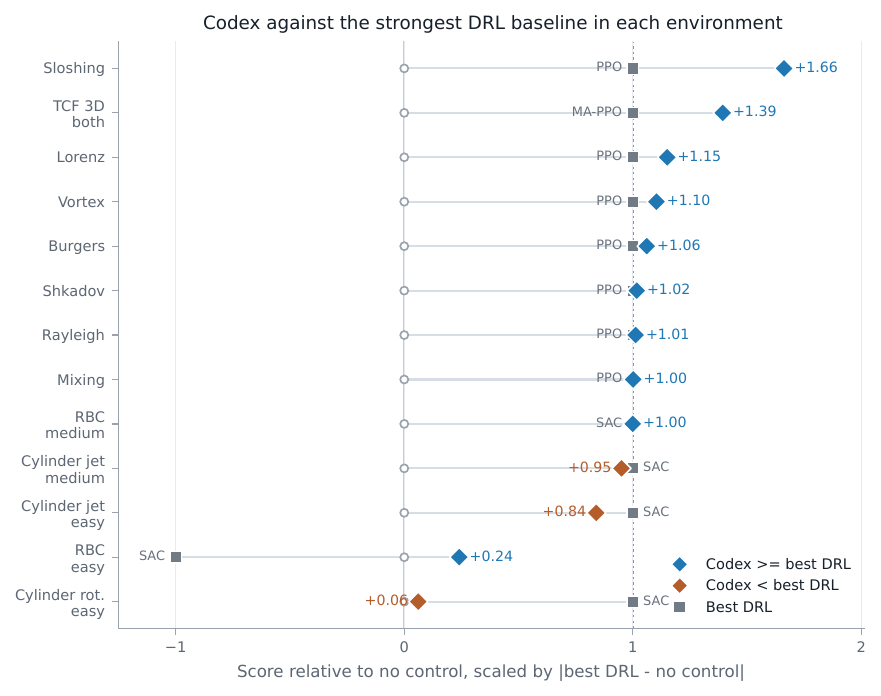}
    \caption{Aggregate comparison between the best coding-agent heuristic and the best DRL baseline on each environment. Scores are normalized by the improvement from the initial DRL control level to the best DRL score, so positive values indicate a heuristic above the best DRL result.}
    \label{fig:codex-vs-best-drl-all-envs}
    \vspace{-1.0em}
\end{figure}

Against the best-performing DRL algorithm on each environment%
\footnote{Among PPO, SAC, and TD3.}, heuristic learning achieves a higher score in 8 environments, matches the best DRL performance in 2 environments, and performs below the strongest baseline in the remaining 3 environments. The complete comparison is reported in \autoref{fig:codex-vs-best-drl-all-envs}. Overall, the proposed approach reaches or exceeds the best DRL performance in 10 of the 13 benchmark problems while producing explicit, fully inspectable controller implementations rather than neural-network policies.

\begin{figure}[b]
\centering
\includegraphics[width=\linewidth]{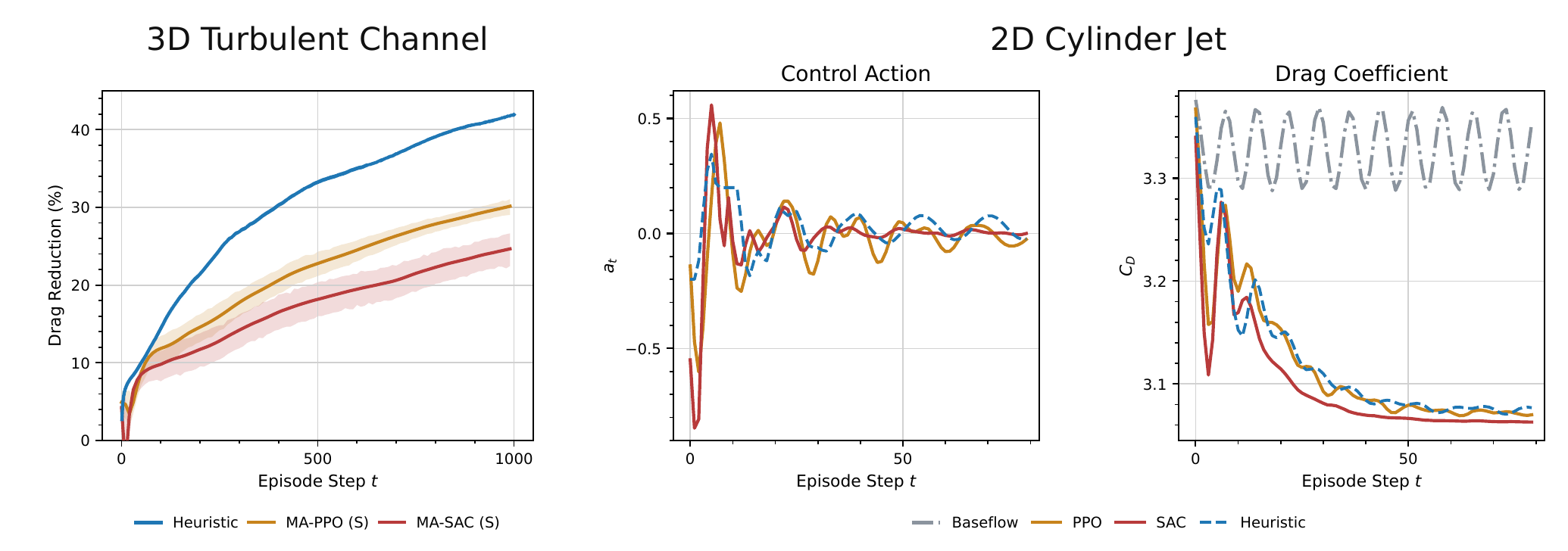}
\caption{Drag reduction obtained by the heuristic controllers in the turbulent-channel and cylinder-control cases. We also present the control actions for the Cylinder case.}
\label{fig:drag-reduction-turbulence-cylinder}
\end{figure}

Among the strongest results are the three-dimensional turbulent-channel configuration \env{TCFSmall3D-both-easy-v0}. In this benchmark, the discovered heuristic not only reaches the performance of the best DRL controller but further improves the normalized score by approximately 50\%. Expressed in physical terms, this corresponds to a drag reduction approaching 40\%, compared with approximately 30\% obtained by the strongest DRL baseline. The corresponding drag histories and control actions are presented in \autoref{fig:drag-reduction-turbulence-cylinder}.

\begin{figure}[b]
\centering
\includegraphics[width=0.86\linewidth]{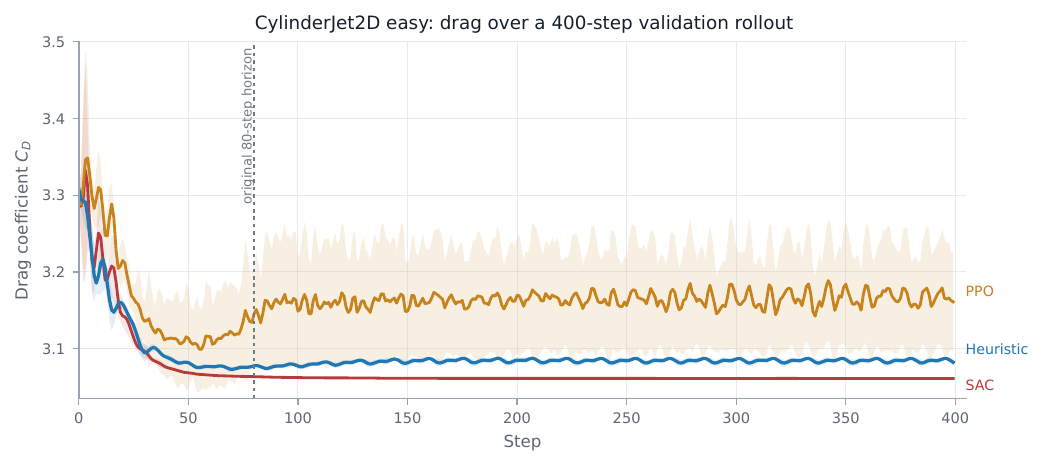}
\caption{Long-horizon 400-step validation drag on \env{CylinderJet2D-easy-v0}. SAC, PPO, and the heuristic are compared with per-step mean traces and uncertainty bands.}
\label{fig:cylinderjet2d-easy-400step-validation-drag}
\end{figure}

The two-dimensional cylinder-jet benchmark introduced by \citet{rabault2019active} provides a complementary perspective. Although the heuristic remains slightly below the best DRL score on the original evaluation horizon, it nevertheless captures the essential feedback mechanism required to substantially reduce drag relative to the uncontrolled flow, as illustrated in \autoref{fig:drag-reduction-turbulence-cylinder}. To further assess the robustness of the discovered controller, we extended the evaluation horizon from the original 80 time steps to 400 time steps without modifying the controller itself. Under these longer rollouts, the PPO controller progressively loses stability shortly after the original training horizon, whereas both the heuristic controller and the SAC policy remain stable throughout the entire simulation. This behavior suggests that the explicit controller has captured a feedback strategy that generalizes beyond the nominal optimization horizon rather than simply exploiting properties specific to the original episode length. The corresponding validation trajectories are shown in \autoref{fig:cylinderjet2d-easy-400step-validation-drag}.

\begin{figure}[!ht]
\centering
\includegraphics[width=\linewidth]{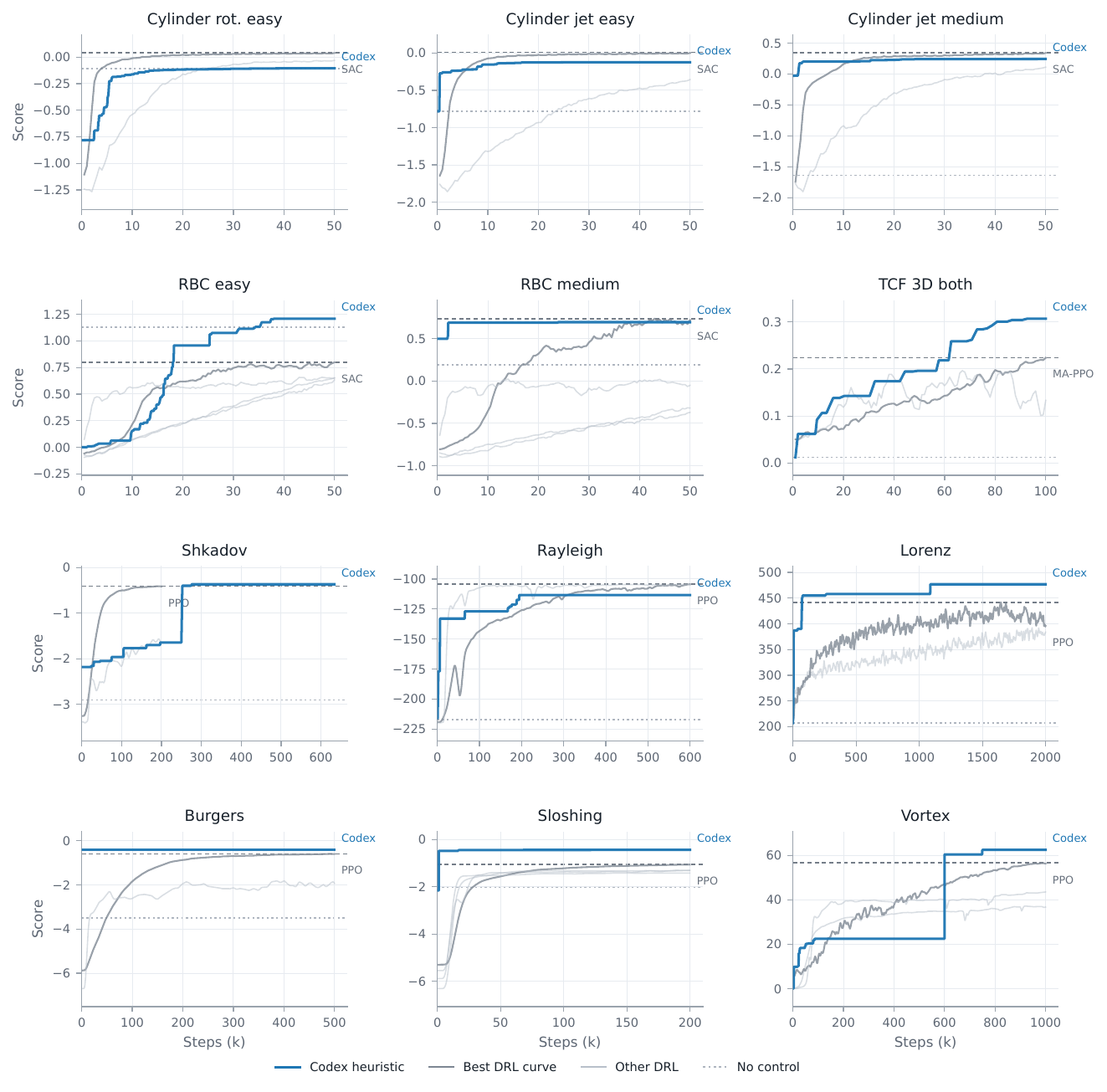}
\caption{Score evolution as a function of consumed environment steps. DRL curves are shown together with the coding-agent heuristic trajectories, with the last available heuristic value extended when the diagnostic used fewer steps than the DRL reference.}
\label{fig:all-results-per-time-step}
\end{figure}

Beyond the final scores, the optimization trajectories provide additional insight into the search dynamics. \autoref{fig:all-results-per-time-step} reports the evolution of the best score as a function of the number of consumed environment interactions for all benchmark problems. Across the suite, heuristic learning consistently exhibits higher sample efficiency during the early stages of optimization, avoiding the prolonged cold-start phase typically observed with DRL. Although small variations are observed during the initial search depending on the explored candidate controllers, repeated runs consistently converge toward similar final heuristic, indicating that the proposed search procedure is robust despite its inherently exploratory nature.

\subsection{Agent comparison}

Beyond evaluating the proposed heuristic-learning framework, the same experimental protocol also provides a controlled benchmark for comparing different coding agents. Since every agent receives exactly the same benchmark environments, source code, prompts, interaction budget, and evaluation procedure, the only varying component is the underlying language model. This allows differences in performance to be attributed directly to the coding capabilities of the agents themselves. Overall, Claude improves upon Gemini by approximately 50\%, while Codex further improves upon Claude by roughly another 50\%. The recently released Fable~5 performs only slightly below Codex.

\paragraph{Scores}

\begin{figure}[!ht]
\centering
\includegraphics[width=0.92\linewidth]{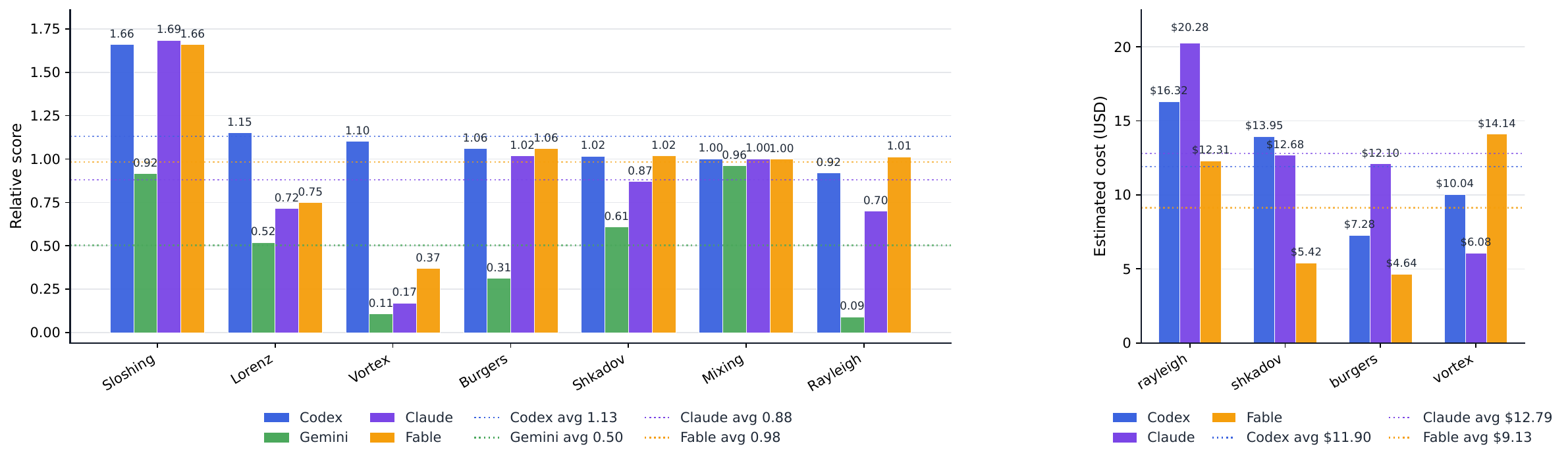}
\caption{Agent comparison on the BEACON environments. Each bar reports the normalized score reached by the corresponding coding agent, ordered from left to right by the Codex score. We also present the cost per environment on the right.}
\label{fig:agent-comparison-codex-gemini-claude}
\end{figure}

We compare four coding agents: Codex, Claude, Gemini, and Fable. All agents operate under the same heuristic-learning harness and are evaluated on the seven BEACON environments. The resulting normalized scores are presented in \autoref{fig:agent-comparison-codex-gemini-claude}. Gemini consistently performs below both Claude and Codex across the benchmark suite. Among the seven environments, Claude outperforms Codex once and matches its performance once, whereas Codex achieves the highest score in the remaining five environments, often by a substantial margin. Averaged over the complete benchmark suite, the normalized scores are \score{1.13} for Codex, \score{0.98} for Fable, \score{0.88} for Claude, and \score{0.50} for Gemini.

\paragraph{Cost analysis}

Besides controller quality, computational cost is an important practical consideration. We therefore compare the different coding agents in terms of token consumption and API cost. On the Shkadov benchmark, Codex used 244 thousand fresh input tokens, 21.0 million cached input tokens, and 74 thousand output tokens, corresponding to a total cost of \$13.95. Under the same conditions, Claude required 114 thousand fresh input tokens, 141 thousand cache-write tokens, 19 million cache-read tokens, and 62 thousand output tokens, for a total cost of \$12.68. Fable exhibited the lowest overall usage, with only 57 thousand cache-write tokens, 2.29 million cache-read tokens, and 40 thousand output tokens, resulting in a total cost of \$5.42.

Across the complete benchmark suite, Codex costs \$11.90 per environment on average, compared with \$12.79 for Claude. When normalized by the obtained benchmark score, Codex requires \$10.53 per score unit, Claude \$14.53, and Fable only \$9.13. Under this metric, Fable~5 provides the most favorable performance-to-cost ratio among the evaluated coding agents. The average computational costs are summarized in \autoref{fig:agent-comparison-codex-gemini-claude}.

\subsection{Other experiments}

\subsubsection{Transfer Learning}

Transfer learning is routinely investigated in reinforcement learning for fluid mechanics by adapting a controller developed for a simpler configuration to a more challenging one. We adopt the same philosophy here. Instead of transferring neural-network parameters, however, the transferred object consists of the explicit controller source code together with its accumulated search ledger, which initializes a new heuristic-learning process.

Three representative transfer scenarios were considered. First, the controller developed for the cylinder-jet benchmark at $Re=100$ was transferred to \env{CylinderJet2D-medium-v0} at $Re=250$. Although the frozen controller performs poorly (\score{-0.6180}), only a limited amount of additional search is required to recover a controller reaching \score{0.2433}. Second, the spatial temperature controller obtained on \env{RBC2D-easy-v0} was transferred to \env{RBC2D-medium-v0}, corresponding to an increase of the Rayleigh number from $8\times10^4$ to $4\times10^5$. The frozen transfer achieves a score of \score{0.4971}, while the adapted controller ultimately reaches \score{0.6937}. Finally, the Shkadov controller originally designed for five actuators was transferred to a ten-actuator configuration, requiring only limited adaptation to recover and eventually exceed the original performance.

\begin{figure}[!ht]
\centering
\includegraphics[width=\linewidth]{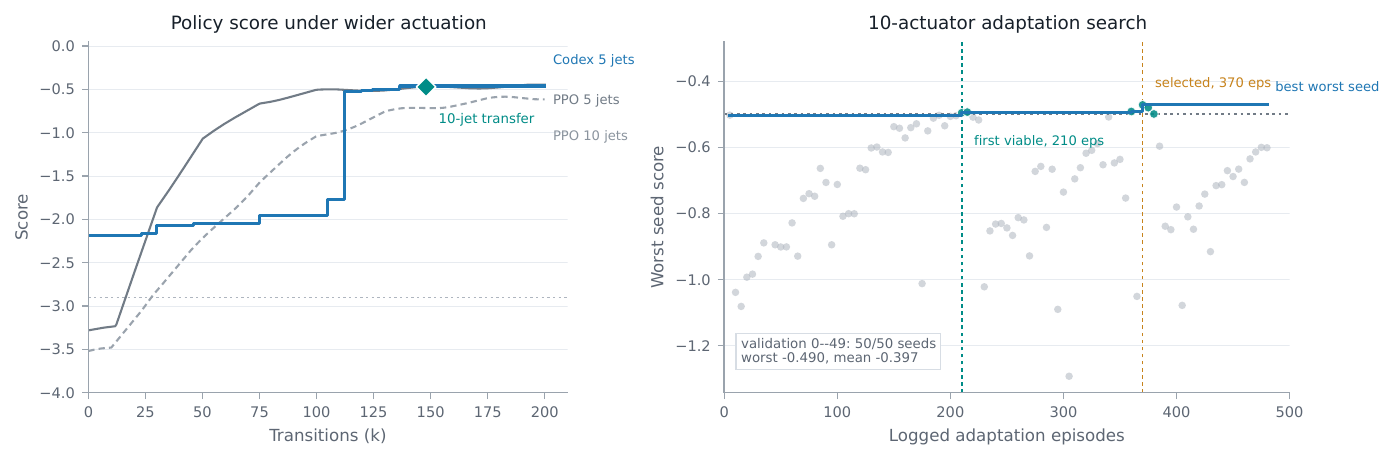}
\caption{Shkadov actuator transfer from 5 to 10 jets. The left panel compares the 5-jet and 10-jet PPO references with the 5-jet heuristic and the selected 10-jet transfer point. The right panel shows the adaptation search that selected the transferred 10-actuator controller.}
\label{fig:shkadov-5-to-10-jets-transfer}
\end{figure}

The transferred Shkadov controller preserves the overall feedback structure. The adapted policy retains the same mean, slope, curvature, and last-point features while modifying only a small number of numerical parameters: the delay changes from \score{11} to \score{8}, the feedback gain from \score{0.3475} to \score{0.28}, the smoothing coefficient from \score{0.84} to \score{0.90}, and the taper from $0.95^j$ to $1.03^j$. These limited modifications make the adaptation process fully interpretable, since the transferred controller differs from the original through only a handful of explicitly identifiable constants. After approximately 500 episodes, the adapted controller also surpasses the strongest DRL baseline. The complete transfer process is illustrated in \autoref{fig:shkadov-5-to-10-jets-transfer}.

\subsubsection{Increasing action and observation dimension}

We next investigate whether the proposed heuristic-learning framework remains effective when the dimensionality of the observations or actions is increased.

\paragraph{Using full mesh fields}

Providing additional information does not necessarily simplify the search problem. To evaluate this effect, we modified \env{CylinderJet2D-easy-v0} so that the controller received full-field diagnostics derived from the public computational mesh, including wake deficit, pressure skew, vorticity skew, divergence proxies, and mesh-based pressure and shear summaries. The best controller exploiting these richer observations achieved a score of \score{-0.3333}, while the selected default controller reached \score{-0.3478}. Although both outperform the uncontrolled flow, neither matches the original public-sensor heuristic (\score{-0.1244}) nor the SAC baseline (\score{0.0005}). During the search, several additional quantities—including recirculation metrics, pressure gradients, and mesh-shear summaries—were explored, but none consistently survived the validation stage. These experiments suggest that increasing the amount of available information does not automatically translate into better explicit controllers and may instead enlarge the search space.

\paragraph{More actuators}

We also investigate a more challenging optimization problem by allowing the dimensionality of the action space itself to vary. Rather than fixing the number of control inputs, the Rayleigh--Bénard benchmark was modified so that the controller first selected the number of actuators before optimizing the corresponding control law. Consequently, the search simultaneously explored controller structure and action-space dimensionality. In the observed optimization trajectories, the coding agent first explored the different actuator configurations before progressively identifying the most promising regions of the search space and refining the corresponding feedback laws. Despite this substantially more difficult optimization problem, the heuristic-learning framework remained capable of identifying effective control strategies.

\subsubsection{Sparse observations}

\begin{figure}[t]
\centering
\includegraphics[width=\linewidth]{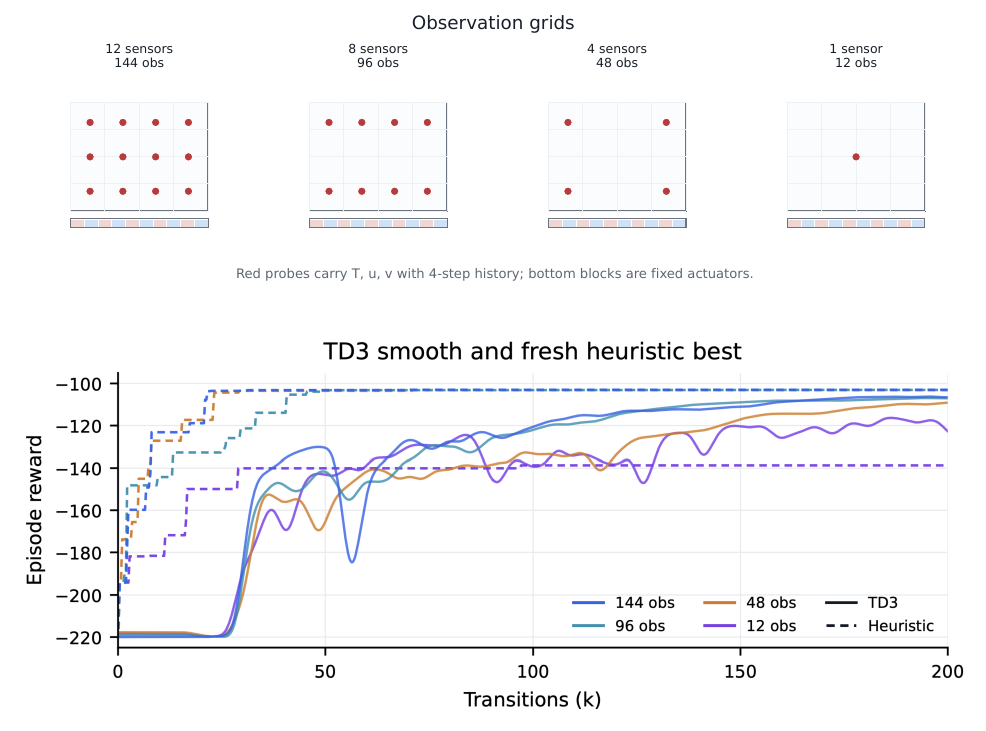}
\caption{Sparse-observation control. The left panel shows the sensor layouts and the bottom actuators; the right panels compare TD3 against the fresh heuristic line searches at 200k transitions.}
\label{fig:rayleigh-sparse-observations-summary}
\end{figure}

Finally, we investigate the opposite situation by progressively reducing the amount of available information. \autoref{fig:rayleigh-sparse-observations-summary} compares heuristic learning with a TD3 controller on the Rayleigh--Bénard benchmark while decreasing the number of observations from 192 to 144, 96, 48, and finally 12. As expected, TD3 gradually deteriorates as observations become sparser, with a pronounced performance drop at only 12 sensors. Heuristic learning exhibits a similar degradation only in this most restrictive configuration, while recovering nearly identical performance to the full-observation case for 144, 96, and 48 observations. Moreover, across all observation levels, the heuristic search reaches competitive controllers substantially earlier than TD3. These experiments indicate that explicit heuristic search remains remarkably robust under partial observations and is capable of identifying effective feedback laws even when only limited information is available.

\section{Policy Anatomy}
\label{sec:policy-anatomy}

The previous section established that heuristic learning is capable of producing controllers that are competitive with state-of-the-art DRL. We now examine the structure of several representative policies in order to better understand the feedback mechanisms discovered during the search. Rather than treating the generated programs as opaque software artifacts, we analyze them as explicit control laws whose individual components can be interpreted physically. Pseudo-code implementations of all reported controllers are provided in \autoref{sec:policies}. We first consider the Shkadov falling-film benchmark, then the three-dimensional turbulent-channel case, and finally compare the learned heuristic with a PPO controller on the cylinder-jet benchmark.

\subsection{Shkadov: delayed shape feedback}

The Shkadov benchmark provides a particularly instructive example because the control problem is inherently convective. The controller observes upstream flow-rate windows, whereas the objective is defined through the downstream film-height error. Consequently, a purely reactive controller is fundamentally suboptimal: by the time an upstream disturbance has been detected and an actuation command is applied, the corresponding wave has already propagated downstream.

The discovered five-actuator controller naturally compensates for this transport delay through a delayed local feedback law. For each actuator-local upstream window $q_j$, it first extracts four physically interpretable features,

\[
\begin{aligned}
\mu_j &= \mathrm{mean}(q_j - 1), \\
s_j &= q_j[-1] - q_j[0],\\
\kappa_j &= q_j[\mathrm{mid}] - \frac{1}{2}(q_j[0] + q_j[-1]),\\
\ell_j &= q_j[-1] - 1.
\end{aligned}
\]

These quantities are then combined into a delayed shape descriptor,

\[
\sigma_j = \mu_j
-0.5847514\,s_j
+0.9217264\,\kappa_j
-0.2278599\,\ell_j .
\]

The resulting control action is

\[
\tilde a_{j,t} =
\operatorname{clip}\left(
-0.3475133\,\sigma_{j,t-11}\,0.95^j + 0.0263648,
-1,1
\right),
\]

followed by a \score{0.05} deadband and exponential smoothing,

\[
a_{j,t}=0.16\,\tilde a_{j,t}+0.84\,a_{j,t-1}.
\]

The numerical coefficients themselves should not be interpreted as a physical model of falling-film dynamics. The important observation is the structure of the discovered controller. The coding agent independently identifies that successful control requires maintaining a compact description of the local wave geometry before applying the corresponding actuation after an appropriate transport delay. The dominant contribution of the curvature term $\kappa_j$ indicates that local wave shape is more informative than the absolute signal itself, while the eleven-step delay approximately synchronizes upstream sensing with downstream actuation. Interestingly, these mechanisms emerge directly from the optimization process without being prescribed by the user.

\subsection{Turbulent channel flow: local wall feedback}

The benchmark \env{TCFSmall3D-both-easy-v0} provides a substantially more challenging interpretability test. Unlike the Shkadov case, the controller no longer receives a small vector of probe measurements but instead observes complete wall fields on both channel walls.

Let the public wall observations be reshaped as $(u_i,v_i)$ over the two wall grids. The controller first normalizes each wall independently,

\[
\hat u_i = \frac{u_i-\overline{u}_{\mathrm{wall}}}
{\mathrm{std}(u_{\mathrm{wall}})+10^{-4}},\qquad
\hat v_i = \frac{v_i-\overline{v}_{\mathrm{wall}}}
{\mathrm{std}(v_{\mathrm{wall}})+10^{-4}}.
\]

It then constructs the local feedback signal

\[
s_i=-0.263671875\,\hat v_i+0.2109375\,\hat u_i,
\]

before applying one nearest-neighbour smoothing pass on each wall, subtracting the mean action independently on each surface, and clipping the resulting commands to the admissible interval \([-0.30,0.30]\).

Ablation studies reveal that every component of this controller contributes to its final performance. Removing the normalization step causes the score to collapse toward \score{0.02}--\score{0.05}, indicating that relative rather than absolute velocity fluctuations are the relevant control variables. Eliminating the positive contribution of the streamwise residual reduces the score to approximately \score{0.10}--\score{0.16}. Likewise, suppressing the smoothing operation, the mean-free projection, or the centering procedure consistently degrades performance, while reversing the sign convention on the upper wall also fails. Finally, replacing the complete wall feedback by a scalar wall-stress signal proved insufficient. Together, these ablations suggest that the controller relies on a genuinely distributed local wall-feedback mechanism rather than on a single global flow indicator.

The remaining policies are detailed in \autoref{sec:policies}.

\subsection{Comparison with PPO based policies}

Finally, we compare the explicit heuristic with policies obtained through conventional deep reinforcement learning on \env{CylinderJet2D-easy-v0}. To this end, we constructed a dataset of 400 observation--action pairs sampled from random trajectories, heuristic rollouts, DRL rollouts, and mixed-policy simulations. This allows the different controllers to be evaluated on identical flow states rather than through aggregate performance alone.

\begin{figure}[!ht]
\centering
\includegraphics[width=0.82\linewidth]{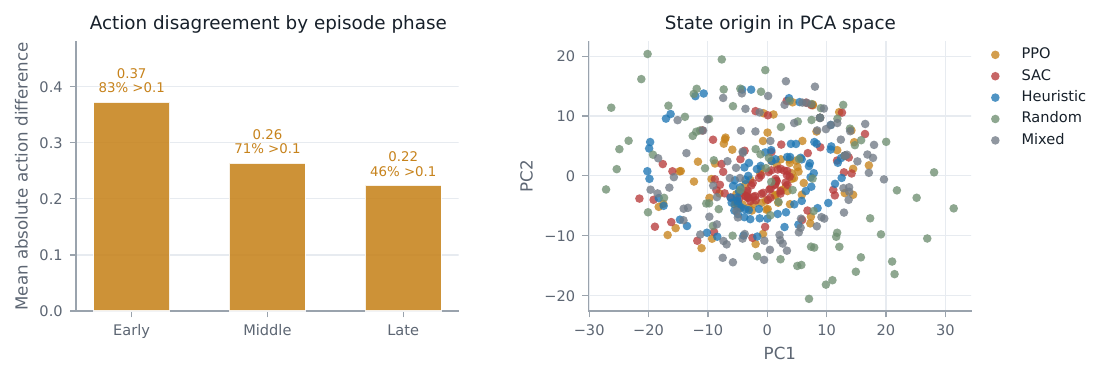}
\caption{Same-state comparison between PPO and the heuristic on \env{CylinderJet2D-easy-v0}. The first panel reports the action gap by episode phase, while the second panel locates the disagreement in the first two principal components of the stored state bank.}
\label{fig:cylinderjet-ppo-vs-heuristic-two-panel}
\end{figure}

Although PPO and the heuristic differ in overall score and long-term stability, their instantaneous control decisions are noticeably more similar than those produced by SAC. The mean absolute action difference between PPO and the heuristic is \score{0.2852}, compared with \score{0.3406} between SAC and the heuristic, while PPO and SAC differ by \score{0.1728}. Furthermore, PPO reproduces the heuristic action within \score{0.05} on \score{21.75\%} of the evaluated states, whereas SAC reaches the same agreement on only \score{14.5\%}. The complete comparison is presented in \autoref{fig:cylinderjet-ppo-vs-heuristic-two-panel}.

The phase-resolved analysis further indicates that the heuristic and PPO become increasingly similar once the controlled wake reaches its quasi-stationary regime, whereas larger discrepancies are observed during the initial transient. The PCA representation provides a complementary interpretation. States originating from random trajectories lie farther from the region explored by the learned controllers, and the action disagreement increases accordingly. Rather than indicating contradictory control strategies, these results suggest that the heuristic and PPO have learned compatible feedback mechanisms on the portion of state space encountered during successful control, while differing primarily in how they extrapolate outside this on-policy region.

\section{Conclusion}

This work investigated whether modern coding agents can serve as practical search procedures for discovering explicit controllers in active flow control. Using identical environments, observations, actions, rewards, and environment-step budgets, the proposed heuristic-learning framework reached or exceeded the strongest available DRL baseline in 10 of the 13 benchmark environments. Unlike conventional reinforcement learning, however, the outcome of the optimization is not a neural-network policy but an explicit executable controller whose feedback structure, gains, delays, and filtering operations remain directly accessible to human inspection.

Beyond the benchmark scores, the experiments demonstrate that explicit controller synthesis provides valuable physical insight. In both the Shkadov and turbulent-channel benchmarks, the discovered heuristics reduce to compact feedback structures that naturally reflect the underlying flow physics. The emergence of delayed shape feedback for convecting falling films and normalized local wall feedback for turbulent channel flow suggests that heuristic learning is capable not only of optimizing performance, but also of recovering meaningful control mechanisms. Furthermore, the transfer experiments show that these controllers can often be adapted to more challenging configurations through the modification of only a few interpretable parameters, rather than requiring the complete retraining of a policy.

The present study also highlights the importance of rigorous evaluation protocols. Increasing the amount of available information does not necessarily simplify controller discovery, while severe observation sparsity ultimately limits performance. More importantly, the validity of the search depends on carefully constraining the coding agent to the public benchmark interface. Without such safeguards, an autonomous agent may exploit implementation details or simulator loopholes rather than discovering admissible control strategies. We therefore believe that future work on agent-based scientific discovery should place increasing emphasis on transparent experimental protocols, reproducible search procedures, and systematic reporting of prompts, search ledgers, rejected candidates, and controller ablations.

More broadly, this work suggests that explicit program synthesis constitutes a credible alternative to parameter
optimization for a class of fluid-control problems. Rather than searching in the high-dimensional space of neural-
network weights, heuristic learning searches directly in the space of executable feedback laws. As coding agents
continue to improve, this paradigm may progressively transform how controllers are designed, analyzed, and transferred,
combining the optimization capabilities of modern AI with the interpretability and physical insight traditionally sought
in computational fluid dynamics.

\section*{Acknowledgments}

We thank OpenAI for support during the experiments.

Funded/Co-funded by the European Union (ERC, CURE, 101045042). Views and opinions
expressed are however those of the author(s) only and do not
necessarily reflect those of the European Union or the European
Research Council. Neither the European Union nor the granting
authority can be held responsible for them.

\bibliographystyle{unsrtnat}
\bibliography{references}

\appendix

\section{Environment Suite}
\label{app:environments}

The benchmark suite contains 13 environments: 6 larger FluidGym flow-control
cases and 7 compact BEACON control cases. \autoref{tab:env-suite-details}
reports the task definition, principal physical parameters, default episode
horizon, and action-step clock used for budget accounting.

\begin{table}[p]
\centering
\scriptsize
\setlength{\tabcolsep}{3.5pt}
\renewcommand{\arraystretch}{1.13}
\caption{Environment characteristics and rollout clocks. Steps are action steps per full default episode. $\Delta t_{\mathrm{env}}$ is FluidGym \texttt{step\_length} or BEACON \texttt{dt\_act}.}
\label{tab:env-suite-details}
\begin{tabularx}{\linewidth}{>{\raggedright\arraybackslash}p{0.22\linewidth} X >{\raggedright\arraybackslash}p{0.26\linewidth} >{\raggedleft\arraybackslash}p{0.055\linewidth} >{\raggedleft\arraybackslash}p{0.10\linewidth}}
\toprule
Environment & Description & Characteristics & Steps & $\Delta t_{\mathrm{env}}$ \\
\midrule
\env{CylinderRot2D-easy-v0} & 2D cylinder wake controlled by wall rotation. & $Re=100$, resolution $24$, $1$ rotation actuator. & 80 & $0.25$ \\
\env{CylinderJet2D-easy-v0} & 2D cylinder wake controlled by synthetic jets. & $Re=100$, resolution $24$, $1$ jet pair. & 80 & $0.25$ \\
\env{CylinderJet2D-medium-v0} & Higher-Reynolds-number jet control for the cylinder wake. & $Re=250$, resolution $32$, $1$ jet pair. & 80 & $0.25$ \\
\env{RBC2D-easy-v0} & Rayleigh-Bénard convection controlled by bottom heaters. & $Ra=8\times 10^4$, $Pr=0.7$, $12$ heaters. & 200 & $1.0$ \\
\env{RBC2D-medium-v0} & Harder Rayleigh-Bénard convection transfer case. & $Ra=4\times 10^5$, $Pr=0.7$, $12$ heaters. & 200 & $1.0$ \\
\env{TCFSmall3D-both-easy-v0} & 3D turbulent channel flow with both-wall actuation. & $Re_\tau=180$, grid $64\times 65\times 64$, both walls. & 1000 & $0.6$ wall units \\
\env{shkadov-v0} & Falling-film height and flow-rate stabilization. & $\delta=0.1$, $5$ jets, $A=5$, $\sigma=5\times 10^{-4}$. & 400 & $0.05$ \\
\env{rayleigh-v0} & 2D convection controlled by bottom temperature segments. & $Ra=10^4$, $Pr=0.71$, $10$ segments, $C=0.75$. & 100 & $2.0$ \\
\env{lorenz-v0} & Forced Lorenz attractor with reward for keeping $x<0$. & $(\sigma,\rho,\beta)=(10,28,8/3)$, $3$ discrete actions. & 500 & $0.05$ \\
\env{burgers-v0} & 1D Burgers inlet-noise damping with point forcing. & $L=2$, $x_p=1$, $A=10$, $\sigma=0.1$, $u^\star=0.5$. & 200 & $0.05$ \\
\env{sloshing-v0} & 1D tank sloshing controlled by cart acceleration. & $L=2.5$, $g=9.81$, $A=5$, $\alpha=5\times 10^{-4}$. & 200 & $0.05$ \\
\env{vortex-v0} & Vortex-induced oscillator with amplitude and phase forcing. & $Re=50$, $m=10$, $\gamma=0.023$, $2$ action coordinates. & 800 & $0.5$ \\
\env{mixing-v0} & 2D passive-scalar mixing driven by wall-shear actions. & $Re=100$, $Pe=10^4$, $4$ discrete wall-shear actions. & 100 & $0.5$ \\
\bottomrule
\end{tabularx}
\end{table}

\section{Heuristics}
\label{sec:policies}

Each algorithm writes the deployed control law, not the search loop that found it.
The operator $\clip(x,\ell,u)$ clips elementwise to $[\ell,u]$.
The operator $\interp(\cdot)$ maps sensor columns to actuator locations.
All controllers use public observations and their own recurrent state unless an open-loop schedule is stated explicitly.

\begin{algorithm}[p]
\caption{\texttt{CylinderRot2D-easy-v0}: PID lift proxy with a small carrier}
\label{alg:cylinder-rot}
\begin{algorithmic}[1]
\Require Observation $o_t$, public info $i_t$, previous lift proxy $\ell_{t-1}$, integral $I_{t-1}$
\State Constants: $A=0.1544$, $b=0.014$, $k_\ell=-1.0$, $k_d=-0.60$, $k_i=0$, $A_p=0.005$, $f=0.055$
\State Split $o_t$ into first and last halves, then $\ell_t \gets \mean(o_t^{\rm first})-\mean(o_t^{\rm last})$
\State $d_t \gets \ell_t-\ell_{t-1}$
\State $I_t \gets \clip(I_{t-1}+\ell_t,-50,50)$
\State $u_t \gets b-k_\ell \ell_t-k_d d_t-k_i I_t+A_p\sin(2\pi f t)$
\State \Return $\clip(u_t,-A,A)$
\end{algorithmic}
\end{algorithm}

\begin{algorithm}[p]
\caption{\texttt{CylinderJet2D-easy-v0}: full-observation lift-proxy feedback}
\label{alg:cylinder-jet-easy}
\begin{algorithmic}[1]
\Require Flattened observation $o_t$, previous proxy $s_{t-1}$
\State Constants: $g=0.19925$, $k_d=1.45$, $b=-0.3933337627$, learned vector $w$, action bounds $[-1,1]$
\State $s_t \gets \clip(b+w^\top o_t,-1,1)$
\State $\Delta s_t \gets 0$ if $t=0$, else $s_t-s_{t-1}$
\State $c_t \gets s_t+k_d\Delta s_t$
\State $u_t \gets -g c_t$
\State \Return $\clip(u_t,-1,1)$
\end{algorithmic}
\end{algorithm}

\begin{algorithm}[p]
\caption{\texttt{CylinderJet2D-medium-v0}: lift PID feedback}
\label{alg:cylinder-jet-medium}
\begin{algorithmic}[1]
\Require Observation $o_t$, public info $i_t$, previous lift proxy $s_{t-1}$, previous difference $\Delta s_{t-1}$, integral $I_{t-1}$
\State Constants: $g=0.18$, $k_d=1.625$, $k_i=0.009$, $I_{\max}=4.0$
\State $s_t \gets$ public lift from $i_t$ if present, otherwise the observation asymmetry proxy
\State $\Delta s_t \gets 0$ if $t=0$, else $s_t-s_{t-1}$
\State $\Delta^2 s_t \gets 0$ if $t=0$, else $\Delta s_t-\Delta s_{t-1}$
\State $I_t \gets \clip(I_{t-1}+s_t,-I_{\max},I_{\max})$
\State $c_t \gets s_t+k_d\Delta s_t+k_i I_t$
\State $u_t \gets -g c_t$
\State \Return $\clip(u_t,-1,1)$
\end{algorithmic}
\end{algorithm}

\begin{algorithm}[p]
\caption{\texttt{RBC2D-easy-v0}: spatial thermal contrast feedback}
\label{alg:rbc2d-easy}
\begin{algorithmic}[1]
\Require Observation vector $o_t$ with 3 channels, action count $n$
\State Constants: gain $g=5.0$, channel weights $(1,0,0)$, contrast power $p=0.5$, smoothing $\alpha=0$
\State Reshape each channel into rows and $n$ actuator blocks
\For{$j=1,\ldots,n$}
  \State $m_j \gets$ mean of channel-0 sensors assigned to actuator $j$
\EndFor
\State $r_j \gets m_j-\mean(m_{1:n})$
\State $s_j \gets \sgn(r_j)|r_j|^p$ for each actuator $j$
\State $u_j \gets -g s_j$
\State $u_t \gets u_t-\mean(u_t)$
\State \Return $\clip(u_t,-1,1)$
\end{algorithmic}
\end{algorithm}

\begin{algorithm}[p]
\caption{\texttt{RBC2D-medium-v0}: smoothed spatial thermal contrast feedback}
\label{alg:rbc2d-medium}
\begin{algorithmic}[1]
\Require Observation vector $o_t$, previous action $u_{t-1}$, action count $n$
\State Constants: gain $g=5.0$, channel weights $(1,0,0)$, contrast power $p=0.5$, smoothing $\alpha=0.3$
\State Compute the local block means $m_{1:n}$ exactly as in Algorithm~\ref{alg:rbc2d-easy}
\State $r_j \gets m_j-\mean(m_{1:n})$
\State $s_j \gets \sgn(r_j)|r_j|^p$
\State $\tilde{u}_j \gets -g s_j$
\State $\tilde{u}_t \gets \tilde{u}_t-\mean(\tilde{u}_t)$
\State $u_t \gets \alpha u_{t-1}+(1-\alpha)\tilde{u}_t$
\State \Return $\clip(u_t,-1,1)$
\end{algorithmic}
\end{algorithm}

\begin{algorithm}[p]
\caption{\texttt{TCFSmall3D-both-easy-v0}: local wall-velocity feedback}
\label{alg:tcf-small-both}
\begin{algorithmic}[1]
\Require Wall observation $o_t$ with local $(u,v)$ features on bottom and top walls
\State Constants: $\alpha=0.263671875$, $\beta_u=0.2109375$, $c=0.3$, one smoothing pass, mean-free per wall
\State Split $o_t$ into bottom and top wall grids, each with local streamwise velocity $u$ and wall-normal velocity $v$
\For{each wall separately}
  \State Center and normalize $u$ and $v$
  \State $s \gets -\alpha v+\beta_u u$
  \State Smooth $s$ once with nearest-neighbor averaging on the wall grid
  \State $s \gets s-\mean(s)$
\EndFor
\State \Return $\clip(s,-c,c)$ on both walls
\end{algorithmic}
\end{algorithm}

\begin{algorithm}[p]
\caption{\texttt{shkadov-v0}: delayed shape feedback with 5 jets}
\label{alg:shkadov}
\begin{algorithmic}[1]
\Require Observation $o_t$ reshaped as 5 actuator-local upstream windows $q_{j,t}$
\State Constants: $w_s=-0.5847514$, $w_c=0.9217264$, $w_l=-0.2278599$, $g=0.3475133$, delay $D=11$
\State Constants: smoothing $\alpha=0.84$, taper $\rho=0.95$, bias $b=0.0263648$, deadband $\delta=0.05$
\For{$j=1,\ldots,5$}
  \State $m_j \gets \mean(q_{j,t}-1)$
  \State $r_j \gets q_{j,t}^{\rm last}-q_{j,t}^{\rm first}$
  \State $c_j \gets q_{j,t}^{\rm mid}-\frac{1}{2}(q_{j,t}^{\rm first}+q_{j,t}^{\rm last})$
  \State $\ell_j \gets q_{j,t}^{\rm last}-1$
  \State $s_{j,t}\gets \clip(m_j+w_s r_j+w_c c_j+w_l\ell_j,-5,5)$
\EndFor
\State Store $s_t$ in a buffer and read $\bar{s}_t=s_{t-D}$, using the oldest available item during startup
\State $\tilde{u}_{j,t}\gets -g\bar{s}_{j,t}\rho^{j-1}+b$
\State $\tilde{u}_{j,t}\gets 0$ when $|\tilde{u}_{j,t}|<\delta$
\State $u_t\gets \alpha u_{t-1}+(1-\alpha)\clip(\tilde{u}_t,-1,1)$
\State \Return $\clip(u_t,-1,1)$
\end{algorithmic}
\end{algorithm}

\begin{algorithm}[p]
\caption{\texttt{rayleigh-v0}: 15-actuator local-plume feedback}
\label{alg:rayleigh}
\begin{algorithmic}[1]
\Require Observation history $o_t$ reshaped into temperature $T$, horizontal velocity $u$, vertical velocity $v$
\State Constants: $n=15$, $g_T=7.64563216$, $g_v=18.0$, $g_{\Delta T}=0.21760451$, $g_{\Delta v}=0.12628452$
\State Constants: spatial smoothing $0.7$, temporal smoothing $0.04918578$, saturation $0.75$, deadband $0.1$
\State Compress the selected middle-$y$ profile over the front history window into columns $T_x$ and $v_x$
\State Interpolate $T_x$ and $v_x$ to the 15 actuator centers
\State Compute front-window finite differences $\Delta T_x$ and $\Delta v_x$, then interpolate them to actuator centers
\State $\tilde{u}_j\gets -(g_TT_j+g_vv_j+g_{\Delta T}\Delta T_j+g_{\Delta v}\Delta v_j)$
\State Smooth $\tilde{u}_{1:n}$ spatially with nearest neighbors
\State $\tilde{u}\gets \tilde{u}-\mean(\tilde{u})$
\State $\tilde{u}_j\gets 0$ when $|\tilde{u}_j|<0.1$
\State $\tilde{u}\gets \clip(\tilde{u},-0.75,0.75)$
\State $u_t\gets 0.04918578\,u_{t-1}+(1-0.04918578)\tilde{u}$
\State \Return $u_t$
\end{algorithmic}
\end{algorithm}

\begin{algorithm}[p]
\caption{\texttt{lorenz-v0}: thresholded recovery controller}
\label{alg:lorenz}
\begin{algorithmic}[1]
\Require Observation $o_t=(x_t,y_t,z_t,\dot{x}_t,\ldots)$ and dwell counter $h$
\State Constants: guard score $\gamma=-5.83480325$, recovery threshold $x_r=-6.08078423$, $w_{\dot{x}}=0.77296558$
\State Constants: guard action $a_g=0$, recovery action $a_r=2$, cruise action $a_c=2$, dwell $D=1$
\If{$h>0$}
  \State $h\gets h-1$
  \State \Return previous action
\EndIf
\State $q_t\gets x_t+w_{\dot{x}}\dot{x}_t$
\If{$x_t>x_r$}
  \State $a_t\gets a_r$
\ElsIf{$q_t>\gamma$}
  \State $a_t\gets a_g$
\Else
  \State $a_t\gets a_c$
\EndIf
\State $h\gets D$
\State \Return $a_t$
\end{algorithmic}
\end{algorithm}

\begin{algorithm}[p]
\caption{\texttt{sloshing-v0}: derivative modal damping}
\label{alg:sloshing}
\begin{algorithmic}[1]
\Require Free-surface observation $o_t$, previous modal projection $p_{t-1}$, previous action $u_{t-1}$
\State Constants: mode $m=1$, gain $g=20.0$, sign $+1$, smoothing $\alpha=0.4$, saturation $1.0$
\State $b_i\gets \sin(m\pi x_i)$ on the observation grid, normalized to unit norm
\State $p_t\gets o_t^\top b$
\State $d_t\gets 0$ if $t=0$, else $p_t-p_{t-1}$
\State $\tilde{u}_t\gets -g d_t$
\State $\tilde{u}_t\gets \clip(\tilde{u}_t,-1,1)$
\State $u_t\gets \alpha u_{t-1}+(1-\alpha)\tilde{u}_t$
\State \Return $u_t$
\end{algorithmic}
\end{algorithm}

\begin{algorithm}[p]
\caption{\texttt{vortex-v0}: bounded piecewise-linear phase schedule}
\label{alg:vortex}
\begin{algorithmic}[1]
\Require Number of action steps $T$, knot vector $\kappa_{1:18}$
\State Constants: amplitude action $a=0.9999516206$
\State Knots:
\Statex \hspace{\algorithmicindent}$\kappa=(-0.2568,-0.3801,-0.4728,-0.5179,-0.5573,-0.5804,$
\Statex \hspace{\algorithmicindent}$-0.6006,-0.6100,-0.6143,-0.6177,-0.6200,-0.6218,$
\Statex \hspace{\algorithmicindent}$-0.6231,-0.6220,-0.6203,-0.6221,-0.6144,-0.5729)$
\State Interpolate $\kappa$ linearly over steps $0,\ldots,T-1$ to get phase sequence $\phi_t$
\For{$t=0,\ldots,T-1$}
  \State $u_t\gets (a,\clip(\phi_t,-1,1))$
  \State Apply $u_t$ to the environment
\EndFor
\end{algorithmic}
\end{algorithm}

\begin{algorithm}[p]
\caption{\texttt{mixing-v0}: alternating shear word}
\label{alg:mixing}
\begin{algorithmic}[1]
\Require Step index $t$
\State Constants: word $(1,2)$ and dwell $D=4$
\State $k\gets \left\lfloor t/D\right\rfloor \bmod 2$
\If{$k=0$}
  \State \Return action $1$
\Else
  \State \Return action $2$
\EndIf
\end{algorithmic}
\end{algorithm}

\end{document}